%% file: main.tex
\documentclass[10pt,twocolumn,letterpaper]{article}

\usepackage[pagenumbers]{cvpr}

\input{preamble}

\definecolor{cvprblue}{rgb}{0.21,0.49,0.74}
\usepackage[pagebackref,breaklinks,colorlinks,allcolors=cvprblue]{hyperref}

\def\confName{CVPR}
\def\confYear{2027}

\title{LoCo-WAM: Test-Time Training as Long Context for World-Action Models}

\author{Sheng Cheng$^{1}$ \qquad Deqian Kong$^{2}$\\
$^{1}$Amazon AGI \qquad $^{2}$University of California, Los Angeles
}

\begin{document}
\maketitle
\input{sec/0_abstract}    
\input{sec/1_intro}
\input{sec/2_related_work}
\input{sec/3_method}
\input{sec/4_exp}

\input{sec/5_conclusion}
\newpage
{
    \small
    \bibliographystyle{ieeenat_fullname}
    \bibliography{main}
}

\input{sec/X_suppl}

\end{document}

%% file: preamble.tex
\usepackage{multirow}
\usepackage{wrapfig}
\usepackage{colortbl}
\newcommand{\std}[1]{{\scriptsize$\pm$#1}}
\newcommand{\pub}[1]{\,{\scriptsize\textcolor{gray}{(#1)}}}
\usepackage{pifont}

\usepackage{amsmath}
\usepackage{amssymb}
\usepackage{algorithm}
\usepackage{algpseudocode}
\usepackage{placeins}

%% file: sec/0_abstract.tex
\begin{abstract}
World-action models guide robot actions by predicting what happens next, but typically condition on a short observation history.
Extending this history increases computation at every decision and can require a separately trained model.
We introduce LoCo-WAM, which learns from the outcomes of its actions through test-time training while keeping the same short history length.
It uses newly observed frames to update a low-rank adapter in the video expert, whose adapted predictions guide subsequent actions.
The updates accumulate within each episode and are reset between episodes, allowing past observations to influence decisions after they leave the input window.
LoCo-WAM thus learns from a longer stream of experience without expert action labels, additional data or meta-training.
Experiments on two world-action models show improved task success across four simulation benchmarks.
On RoboCasa GR-1, a model with 2.4\,s of history matches one trained with 9.6\,s, while a model with 4.8\,s exceeds it by 4.8 percentage points.
Our analysis shows that the updates improve prediction of upcoming frames, affect the action mainly through adaptation of the imagined future, and have an increasing effect on actions as the episode unfolds.
\end{abstract}

%% file: sec/1_intro.tex
\section{Introduction}
\label{sec:intro}

\begin{figure}[t]
\centering
\includegraphics[width=\columnwidth]{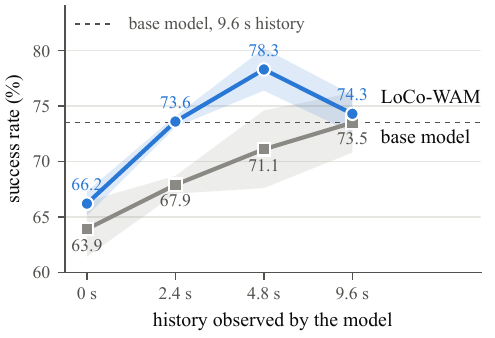}
\caption{\textbf{Test-time training as long context.} Success rate of Long-WAM on RoboCasa GR-1 for each history length, without and with LoCo-WAM. With LoCo-WAM, a model with 2.4\,s of history matches the base model with 9.6\,s, and a model with 4.8\,s of history exceeds it.}
\label{fig:teaser}
\end{figure}

Human behavior combines knowledge acquired over time with rapid learning from new experience, as described by complementary learning systems theory~\cite{mcclelland1995complementary}.
In motor control, discrepancies between predicted and observed sensory outcomes can drive adaptation during action~\cite{tseng2007sensory}.
These findings motivate learning during robot execution, where the outcomes of actions provide feedback for the predictions that guide subsequent behavior~\cite{noh2025latent}.

World-action models~\cite{unipi,uva,cosmospolicy,lingbotva,longwam} provide a natural setting for such learning.
They predict future frames together with actions, drawing on video pretraining to learn how scenes evolve.
In the models we study, the imagined future guides action generation.
After the robot acts, the observed frames provide targets for the model's own video loss.
As interactions within a task share objects, surroundings and dynamics, learning from these frames may improve predictions of what happens next.
The updated predictions can then guide subsequent actions.

This learning allows experience to influence behavior beyond a fixed observation window.
Most world-action models condition on a short history, which limits the past observations available at each decision.
Extending the history~\cite{longwam,lingbotva} increases computation and can require a separately trained model.
Selecting or compressing the history requires components that are trained together with the policy~\cite{hilowam,memorywam,dimwam}.
Updating the model on the observed frames provides another way to use the past observations.
These updates are a form of test-time training, a family of methods that update a model on each test input with a self-supervised loss~\cite{ttt_sun}, which has been extended to language and video generation models~\cite{templora,slowfastvgen,ttte2e,tttlayers,oneminute}.

We introduce LoCo-WAM, a test-time training method for world-action models, which trains the video expert during execution with its video loss on the observed frames.
At every decision, LoCo-WAM takes one gradient step on a low-rank adapter~\cite{lora} of the video expert to predict the most recently observed frames from the history that preceded them.
It keeps the action expert unchanged and resets the adapter at the start of every episode.
Since the adapter accumulates these updates over the episode, the video expert is trained on frames that have already left its history.
The action expert then generates the next action from the imagined future and the features of the adapted video expert.
LoCo-WAM thereby requires no labels, rewards, additional data, new layers or meta-training, keeps the history length of the model, and can run its update while the robot executes the current action chunk.

We evaluate LoCo-WAM on two world-action models, Long-WAM~\cite{longwam} and LingBot-VA~\cite{lingbotva}, and four simulation benchmarks: RoboCasa GR-1~\cite{gr00tn1}, RoboCasa365~\cite{robocasa365}, LIBERO-Mem~\cite{liberomem} and RoboTwin 2.0~\cite{robotwin2}.
LoCo-WAM improves the base model on all four benchmarks.
On RoboCasa GR-1, Long-WAM with 2.4\,s of history reaches 73.6\% with LoCo-WAM, on par with the base model trained with 9.6\,s of history, at 0.76\,s per decision compared with 0.67\,s and without training a separate model.
With 4.8\,s of history, it reaches 78.3\%, 4.8 points above the 9.6\,s model (\cref{fig:teaser}).
On LIBERO-Mem, LingBot-VA with an 8-chunk history matches its default 30-chunk history.
On both benchmarks, LoCo-WAM also outperforms test-time methods for robot policies, including AdaWorldPolicy~\cite{adaworldpolicy}, which adapts the world model across the episodes of a task, and RoboTTT~\cite{robottt}, which meta-trains fast-weight layers.

Our analysis on RoboCasa GR-1 examines what an update of the video expert learns from the observed frames, how it reaches the action, and how its effect develops over the episode.
The update improves the prediction of upcoming frames before they are used for training, with larger reductions in video loss for shorter histories.
Using the model's own predictions as targets in place of the observed frames removes the gain in success rate.
The update affects the action mainly through the imagined future, since applying it only when the video expert predicts this future recovers most of the gain.
We further show that updating only the video expert performs better than also updating the action expert with the action loss.
Over the episode, the change of the action accumulates, and stopping the updates early removes most of the gain.

In summary, our contributions are:
\begin{itemize}
\item We introduce LoCo-WAM, test-time training as long context for world-action models: the video expert is trained during execution with its video loss on the observed frames, without labels, additional data or meta-training.
\item On two world-action models and four simulation benchmarks, LoCo-WAM improves the base model on all four benchmarks, enables a model with a short history to match or exceed the base model with a longer history, and outperforms prior test-time methods for robot policies.
\item We analyze how the update of the video expert reaches the action: it improves the prediction of upcoming frames from the observed frames, affects the action mainly through the imagined future, and the change of the action it causes accumulates over the episode.
\end{itemize}

%% file: sec/2_related_work.tex
\section{Related Work}
\label{sec:related}

\paragraph{World-action models.}
World-action models predict future frames together with actions, which enables them to benefit from large-scale video pretraining and to use the predicted future to choose actions~\cite{unipi,kong2026plan,kong2024latent,noh2025latent,cheng2025latent}.
Examples include UniPi~\cite{unipi}, GR-1~\cite{gr1bytedance}, VPP~\cite{vpp}, UVA~\cite{uva}, UWM~\cite{uwm}, Cosmos Policy~\cite{cosmospolicy}, OpenWAM~\cite{openwam}, Fast-WAM~\cite{fastwam} and LingBot-VA~\cite{lingbotva}, many of which pair a pretrained video model that predicts the future frames with an action model that generates the actions from this prediction.

Many manipulation tasks are long-horizon, and the information needed to act, such as which steps have already been completed, is often no longer visible in the current observation, so a policy has to rely on its history.
Most world-action models, however, condition on a single frame or a short history.
Some methods train the model with longer histories, with a separate model for each history length in Long-WAM~\cite{longwam} and with long video-action sequences in LingBot-VA~\cite{lingbotva}.
Others select or compress the history, with a downsampled history stream in HiLo-WAM~\cite{hilowam}, anchor frames and gist tokens in MemoryWAM~\cite{memorywam}, event memory banks in DiM-WAM~\cite{dimwam}, and the first observation next to the recent history in ReCAP~\cite{recap}.
LoCo-WAM keeps the short-history model and trains it at test time on the observed frames, which enables it to reach the performance of a longer history.

\paragraph{Test-time training.}
Test-time training updates a model on each test input with a self-supervised loss and was introduced to handle distribution shift~\cite{ttt_sun,tent,tttmae,tttvideo}.
For long context, dynamic evaluation~\cite{dynamiceval}, Temp-LoRA~\cite{templora}, TTT-E2E~\cite{ttte2e} and SlowFast-VGen~\cite{slowfastvgen} update the weights of the model, or a temporary LoRA, on the context during inference, and TTT layers~\cite{tttlayers}, Titans~\cite{titans}, LaCT~\cite{lact} and One-Minute Video Generation~\cite{oneminute} build such updates into the architecture as fast-weight layers.
LoCo-WAM belongs to the first group and trains a low-rank adapter of a world-action model with its own video loss.
In a world-action model, the self-supervised task, video prediction, shares the video expert with the main task, the prediction of the action.

In robotics, PAD~\cite{pad} adapts a visuomotor policy during deployment with a self-supervised loss.
AdaWorldPolicy~\cite{adaworldpolicy}, WorldAgen~\cite{worldagen} and T$^3$VF~\cite{t3vf} adapt a world model with its prediction loss, mainly to handle distribution shift, and SALT~\cite{salt} adapts the action side under visual disruptions.
RoboTTT~\cite{robottt} and WAM-TTT~\cite{wamttt} meta-train fast-weight layers, to extend the context of a policy and to learn from human videos, respectively.
Test-time reinforcement learning~\cite{evolvevla,ttvla} updates a policy with a reward signal.
LoCo-WAM updates the video expert with its own video loss on the observed frames and resets the adapter after every episode, without meta-training or rewards.

%% file: sec/3_method.tex
\section{Method}
\label{sec:method}

This section presents LoCo-WAM.
We first describe how a world-action model acts from a fixed history (\cref{sec:wam}).
We then introduce LoCo-WAM, which trains a world-action model at test time on the current episode with its pretraining losses (\cref{sec:ttt}).
Finally, we describe the setting used in our experiments, which updates only the video expert, and how the update reaches the action (\cref{sec:readout}).

\subsection{World-action models with a fixed history}
\label{sec:wam}

A world-action model consists of a video expert with parameters $\Phi_v$ and an action expert with parameters $\Phi_a$.
At decision $t$, it receives the recent observation history $H_t$ of a fixed length $P$, together with the language instruction and the proprioceptive state.
The video expert first predicts the latents of the upcoming frames from the history, independently of the action tokens of the current chunk, and the action expert then samples the next action chunk conditioned on the history and this prediction:
\begin{align}
\hat{z}_t &= \operatorname{Rollout}_{\Phi_v}\big(\epsilon^v \mid H_t\big), \label{eq:rollout} \\
a_t &= \operatorname{Denoise}_{\Phi_a}\big(\epsilon^a \mid H_t, \hat{z}_t;\, \Phi_v\big) , \label{eq:denoise}
\end{align}
where $\operatorname{Rollout}$ and $\operatorname{Denoise}$ are multi-step denoising processes, $\epsilon^v$ and $\epsilon^a$ are independent Gaussian noise, and $\hat z_t$ denotes the predicted latents of the imagined future.
In this predict-then-act structure, the action expert acts as an inverse dynamics model that infers the action from the observation history and the predicted future.
The video expert thus not only produces the imagined future in $\operatorname{Rollout}$ but also computes the features of the history and the imagined future that the action expert reads in $\operatorname{Denoise}$.
The executed action chunk then produces the next observations, which enter the history $H_{t+1}$ of the next decision.

Both experts are trained with flow matching.
For a noise level $\sigma \in [0, 1]$, distinct from the decision index $t$, a clean target $x$ is mixed with Gaussian noise $\epsilon$ as $x^{\sigma} = (1-\sigma)\, x + \sigma\, \epsilon$, and the model predicts the velocity $\epsilon - x$:
\begin{align}
\mathcal{L}_{\text{video}}(\Phi_v) &= \mathbb{E} \big\| v_{\Phi_v}(z^{\sigma}, \sigma \mid H) - (\epsilon - z) \big\|_2^2 , \label{eq:video_loss} \\
\mathcal{L}_{\text{action}}(\Phi_a) &= \mathbb{E} \big\| u_{\Phi_a}(a^{\sigma}, \sigma \mid H, \tilde{z}; \Phi_v) - (\epsilon - a) \big\|_2^2 , \label{eq:action_loss}
\end{align}
where $H$ is the history of a training window, $z$ the ground-truth latents of the frames that follow it, and $\tilde{z}$ the corresponding noisy latents.
The total objective is $\mathcal{L} = \mathcal{L}_{\text{video}} + \lambda\, \mathcal{L}_{\text{action}}$.
Because the history length $P$ is fixed during training, information older than $P$ frames is not available to the model.
Using a longer history requires training a model with a longer context, which also processes more tokens at every decision.

\subsection{Test-time training of world-action models}
\label{sec:ttt}

LoCo-WAM trains a world-action model at test time on the current episode, without changing its pretrained weights or its history length.
LoCo-WAM parameterizes the test-time update with a low-rank adapter~\cite{lora} in each expert, so that each adapted linear layer computes
\begin{equation}
W' = W + \frac{\alpha}{r} B A ,
\label{eq:lora}
\end{equation}
where $W$ is the frozen pretrained weight, $A$ and $B$ are matrices of rank $r$, $\alpha$ is a scaling factor, and $B$ is initialized to zero, so that the adapted model equals the pretrained model at the start of every episode.

We denote the adapters of the video and action experts by $\Theta_v$ and $\Theta_a$.
At every decision $t$, LoCo-WAM forms a training sample from the most recent window of the episode whose future frames have all been observed. Its input is the history $H$ preceding the window, and its targets are the latents $z$ of the frames observed in it and the action chunk $a$ executed at its start.
Each loss updates the adapter of the expert it supervises:
\begin{align}
\Theta_v &\leftarrow \Theta_v - \eta_v\, \operatorname{clip}\big(\nabla_{\Theta_v} \mathcal{L}_{\text{video}}(\Phi_v + \Theta_v)\big) , \label{eq:update} \\
\Theta_a &\leftarrow \Theta_a - \eta_a\, \operatorname{clip}\big(\nabla_{\Theta_a} \mathcal{L}_{\text{action}}(\Phi_a + \Theta_a)\big) , \label{eq:update_action}
\end{align}
where $\operatorname{clip}$ denotes gradient clipping, and at every decision \cref{eq:update} and \cref{eq:update_action} are applied for $S_v$ and $S_a$ steps, with learning rates and step counts chosen independently for the two experts.
The adapters persist across the decisions of an episode and are reset at the beginning of each episode.
LoCo-WAM therefore requires no labels, no additional data and no meta-training, since its targets are the observations and actions of the episode and its losses are those of pretraining.

\subsection{From the video update to the action}
\label{sec:readout}

In our design, LoCo-WAM updates only the video expert with \cref{eq:update}.
The targets of the video loss are frames produced by the environment, which the model could not observe when it made its prediction, and updating the video expert thus corrects its prediction with information from the episode.
\Cref{alg:loco} summarizes the procedure.

\begin{algorithm}[t]
\caption{LoCo-WAM}
\label{alg:loco}
\renewcommand{\algorithmicrequire}{\textbf{Input:}}
\begin{algorithmic}[1]
\Require video expert $\Phi_v$, action expert $\Phi_a$, history length $P$, learning rate $\eta_v$, steps $S_v$
\State initialize $\Theta_v$ with $B = 0$ \Comment{every episode}
\For{each decision $t = 0, 1, 2, \dots$}
  \State $H_t \gets$ the last $P$ observations
  \If{a new chunk is fully observed}
    \For{$s = 1, \dots, S_v$}
      \State $\Theta_v \gets \Theta_v - \eta_v\, \operatorname{clip}\big(\nabla_{\Theta_v} \mathcal{L}_{\text{video}}(\Phi_v + \Theta_v)\big)$ \Comment{\cref{eq:update}}
    \EndFor
  \EndIf
  \State $\Theta_t \gets \Theta_v$
  \State $\hat{z}_t \gets \operatorname{Rollout}_{\Phi_v + \Theta_t}(\epsilon^v \mid H_t)$ \Comment{\cref{eq:loco_rollout}}
  \State $a_t \gets \operatorname{Denoise}_{\Phi_a}(\epsilon^a \mid H_t, \hat{z}_t; \Phi_v + \Theta_t)$ \Comment{\cref{eq:loco_denoise}}
  \State execute $a_t$
  \State append the resulting observations to the history
\EndFor
\end{algorithmic}
\end{algorithm}

At inference, the adapted video expert $\Phi_v + \Theta_t$, with $\Theta_t$ the adapter after the update at decision $t$, replaces $\Phi_v$ in both \cref{eq:rollout} and \cref{eq:denoise}, while the action expert is unchanged:
\begin{align}
\hat{z}_t &= \operatorname{Rollout}_{\Phi_v + \Theta_t}\big(\epsilon^v \mid H_t\big), \label{eq:loco_rollout} \\
a_t &= \operatorname{Denoise}_{\Phi_a}\big(\epsilon^a \mid H_t, \hat{z}_t;\, \Phi_v + \Theta_t\big) . \label{eq:loco_denoise}
\end{align}
\Cref{eq:loco_rollout} describes the effect of the update on the video. Since $\Theta_t$ is trained on the frames observed in the episode, it adapts the imagined future to the dynamics of this episode.
\Cref{eq:loco_denoise} describes its effect on the action, which arises in two ways, through the imagined future $\hat{z}_t$ and through the features of the history and the imagined future that the adapted video expert computes in $\operatorname{Denoise}$.
\Cref{sec:analysis} analyzes both effects empirically.

\textbf{Asynchronous update.} In the synchronous form, the $S_v$ update steps of each decision, each requiring one forward and one backward pass through the adapted layers of the video expert, add to the latency from observation to action.
To reduce this latency, the update can run asynchronously.
Since the robot executes the current action chunk before the next decision, the update runs concurrently with execution, and $\Theta_t$ in \cref{eq:loco_rollout,eq:loco_denoise} is replaced by $\Theta_{t-1}$, the adapter obtained after all $S_v$ steps of the previous decision.
This requires the $S_v$ steps to finish within the execution of one action chunk.
Unless marked as asynchronous, results use the synchronous update.

%% file: sec/4_exp.tex
\section{Experiments}

\subsection{Setup}
\label{sec:setup}

\paragraph{Evaluation benchmarks.}
We apply LoCo-WAM to Long-WAM~\cite{longwam} and LingBot-VA~\cite{lingbotva}, two world-action models that condition on a fixed history.
We evaluate each base model on two closed-loop simulation benchmarks under the official protocols, run each setting three times, and report the mean and standard deviation of the success rate in percent.

\textbf{RoboCasa GR-1.}
RoboCasa GR-1~\cite{gr00tn1} consists of 24 tabletop tasks for a GR-1 humanoid, 18 pick-and-place and 6 articulated tasks, whose demonstrations last 10 to 18\,s on average.
Long-WAM provides models trained with 0, 2.4, 4.8 and 9.6\,s of history, which serve as our short- and long-context models, and each evaluation runs 10 trials per task.

\textbf{RoboCasa365.}
RoboCasa365~\cite{robocasa365} contains atomic and composite household tasks in simulated kitchens.
Following Long-WAM, we evaluate its released model with 2.4\,s of history on the 50 target tasks with 50 trials per task.

\textbf{LIBERO-Mem.}
LIBERO-Mem~\cite{liberomem} consists of 10 tabletop tasks in which a Franka arm repeatedly manipulates the same objects, so that the progress of the task cannot be inferred from the current observation.
We fine-tune LingBot-VA on its demonstrations and evaluate it on 20 held-out initial states per task with an 8-chunk history, using the default 30-chunk history as the long-history reference, and report the success rate and the subgoal completion rate.

\textbf{RoboTwin 2.0.}
RoboTwin 2.0~\cite{robotwin2} is a bimanual benchmark with 50 tasks, on which LingBot-VA with its default 72-chunk history already reaches 91.5\% average success.
We therefore select the 8 tasks with the lowest success rate and split them into the 4 tasks on which the 72-chunk history clearly outperforms a 16-chunk history in the full-benchmark evaluation and the remaining 4, with 100 trials per task in scenes without domain randomization.
\input{tables/tab_gr1}
\paragraph{Baselines.}
We compare LoCo-WAM with three kinds of baselines, namely prior methods with the results reported in the literature, the base model without test-time updates, and test-time methods that we reimplement on the same base models.
The base model is evaluated both with the same history as LoCo-WAM and with a longer history, up to 9.6\,s for Long-WAM and 30 and 72 chunks for LingBot-VA on LIBERO-Mem and RoboTwin 2.0.
\begin{itemize}
\item AdaWorldPolicy~\cite{adaworldpolicy} updates a LoRA with the prediction loss of the world model and accumulates the adaptation across the episodes of a task.
\item RoboTTT~\cite{robottt} adds fast-weight layers to the action expert, meta-trains them on the benchmark demonstrations and updates them at every decision. We also evaluate it without the memory update.
\item Future Consensus~\cite{futureconsensus} samples several candidate futures at every decision and executes the one closest to their mean, which increases the inference cost by the number of candidates.
\item ReCAP~\cite{recap} keeps the first observation as an anchor next to the most recent history and thus extends the input of the model.
\end{itemize}

\paragraph{Experiment setting.}
Unless stated otherwise, LoCo-WAM adds a LoRA of rank 32 to the self-attention and feed-forward layers of the video expert and takes one update step per decision, $S_v = 1$, with a learning rate that depends on the base model and the benchmark.

\subsection{Main results}

\paragraph{Long-WAM on RoboCasa GR-1.}
\Cref{tab:gr1} shows that LoCo-WAM enables Long-WAM with a short history to reach the performance of a longer history.
It improves the base model by 2.3, 5.7 and 7.2 points with 0, 2.4 and 4.8\,s of history, and by 0.8 points with 9.6\,s.
With 2.4\,s of history, it matches the 9.6\,s base model, 73.6\% compared with 73.5\%, and with 4.8\,s of history, it reaches 78.3\%, 4.8 points above the 9.6\,s base model and the best result among all prior methods trained on the same data.

With 2.4\,s of history, LoCo-WAM also improves the base model more than the other test-time methods, by 5.7 points compared with 2.5 points for AdaWorldPolicy and 3.8 and 3.9 points for RoboTTT with and without memory.

On the 18 pick-and-place tasks, LoCo-WAM improves the 4.8\,s model by 8.5 points to 83.3\%, above the 9.6\,s base model at 79.8\% and all prior methods trained on the same data.
On the 6 articulated tasks, a longer history does not help the base model, which drops from 60.0\% at 4.8\,s to 54.5\% at 9.6\,s, whereas LoCo-WAM improves it at every history length, by 7.3, 0.5, 3.4 and 8.8 points.

\input{tables/tab_rc365}
\paragraph{Long-WAM on RoboCasa365.}
\Cref{tab:rc365} shows that LoCo-WAM also improves Long-WAM on RoboCasa365.
With 2.4\,s of history, it improves the base model by 1.4 points, with gains on all three task categories, by 1.2 points on atomic tasks, 1.3 points on seen composite tasks and 1.6 points on unseen composite tasks, whereas AdaWorldPolicy leaves the success rate unchanged.

\input{tables/tab_liberomem}
\paragraph{LingBot-VA on LIBERO-Mem.}
\Cref{tab:liberomem} shows that LoCo-WAM closes the gap between a short and a long history for LingBot-VA.
With an 8-chunk history, it improves the base model by 4.7 points in success rate and 3.1 points in subgoal completion, matching the default 30-chunk history on both metrics.
LoCo-WAM also improves the base model more than the other test-time methods, by 4.7 points in success rate compared with at most 2.8 points for Future Consensus.

\input{tables/tab_robotwin}
\paragraph{LingBot-VA on RoboTwin 2.0.}
\Cref{tab:robotwin} shows that LoCo-WAM also improves LingBot-VA on RoboTwin 2.0.
On the 8 tasks with the lowest success rate, it improves LingBot-VA with the default 72-chunk history by 1.1 points, compared with 0.2 points for AdaWorldPolicy.
With a 16-chunk history, it improves the base model from 51.8\% to 53.3\% on the 4 tasks that benefit most from a long history, compared with 54.2\% for the 72-chunk history.
In contrast, on the other 4 tasks, where the two histories perform alike, it leaves the 16-chunk model unchanged at 77.3\%.
The gain of LoCo-WAM with a short history is thus concentrated on the tasks that benefit from a long history.

\subsection{Compute cost}
LoCo-WAM substitutes for a longer history at a moderate cost in latency.
With the synchronous update, a decision of LoCo-WAM with 2.4\,s of history takes 0.76\,s, compared with 0.67\,s for the base model with 9.6\,s of history, whose success rate it matches, as the latency column of \cref{tab:gr1} shows.
The asynchronous update of \cref{sec:readout} removes the update from the time between observation and action, since it runs while the robot executes the current action chunk.
With it, a decision takes 0.62\,s with 2.4\,s of history and 0.64\,s with 4.8\,s, less than the 0.67\,s of the 9.6\,s base model, at success rates of 74.9\% and 76.9\%, the latter 3.4 points above the 9.6\,s base model.
The update still runs at every decision, so the compute per decision remains 0.81\,s.
LoCo-WAM increases the peak GPU memory in \cref{tab:gr1} from 13.7 to 14.6\,GB with 2.4\,s of history, for the adapter and its backward pass, compared with 14.0\,GB for the base model with 9.6\,s and 15.4\,GB for AdaWorldPolicy.
LoCo-WAM also needs no additional offline training, whereas the 9.6\,s model is a separately trained model, and RoboTTT requires about 136 GPU-hours of meta-training and takes 1.06\,s per decision.

\subsection{Ablations}
\input{tables/tab_ablation_main}
\input{tables/tab_side}

We ablate LoCo-WAM on Long-WAM on RoboCasa GR-1 in two parts, the configuration of the adapter and its update in \cref{tab:ablation_main}, and the loss used for the update in \cref{tab:side}.
With 4.8\,s of history, the default configuration of \cref{sec:setup} reaches 78.3\%, and \cref{tab:ablation_main} changes one of its components at a time.
A sufficiently large learning rate $\eta_v$ is needed, since a three times smaller one lowers the success rate to 72.8\%, whereas a larger one keeps it at 77.7\%.
More update steps on the same training sample do not help, however, as $S_v = 2$ and $S_v = 4$ lower the success rate to 73.6\% and 73.9\%.
The adapter needs both the self-attention and the feed-forward layers of the video expert, since either part alone loses most of the gain, with 71.4\% and 72.8\%.
The rank shows no clear trend, and the default $r = 32$ performs best.
All variants in \cref{tab:ablation_main} remain at or above the base model at 71.1\%.

\Cref{tab:side} updates with the video loss, as in LoCo-WAM, with the action loss, or with both, each training an adapter on the expert it supervises with its own learning rate.
Every update improves the base model, but the video loss alone performs best with both history lengths, and adding the action loss to it lowers the success rate from 78.3\% to 73.7\% with 4.8\,s of history.
The video update changes the action in two ways in \cref{eq:loco_denoise}, through the imagined future $\hat{z}_t$ and through the features of the history and the imagined future that the adapted video expert computes, and it thus adapts the action without training the action expert.
The action update, in contrast, changes only the action expert, and it affects later decisions only through the executed actions, which change the observed frames in their history.

\subsection{Analysis of LoCo-WAM}
\label{sec:analysis}
\begin{figure}[t]
\centering
\includegraphics[width=\columnwidth]{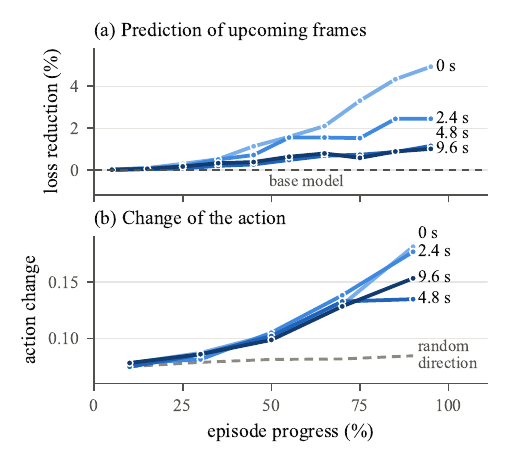}
\caption{\textbf{Prediction and action during an episode.} (a) Relative reduction of the video loss on the next chunk. (b) L2 distance between the action chunks with and without the update, and with a random direction of the same norm, averaged over the history lengths. Long-WAM on RoboCasa GR-1.}
\label{fig:mechanism}
\end{figure}
\begin{figure*}[t]
\centering
\includegraphics[width=\textwidth]{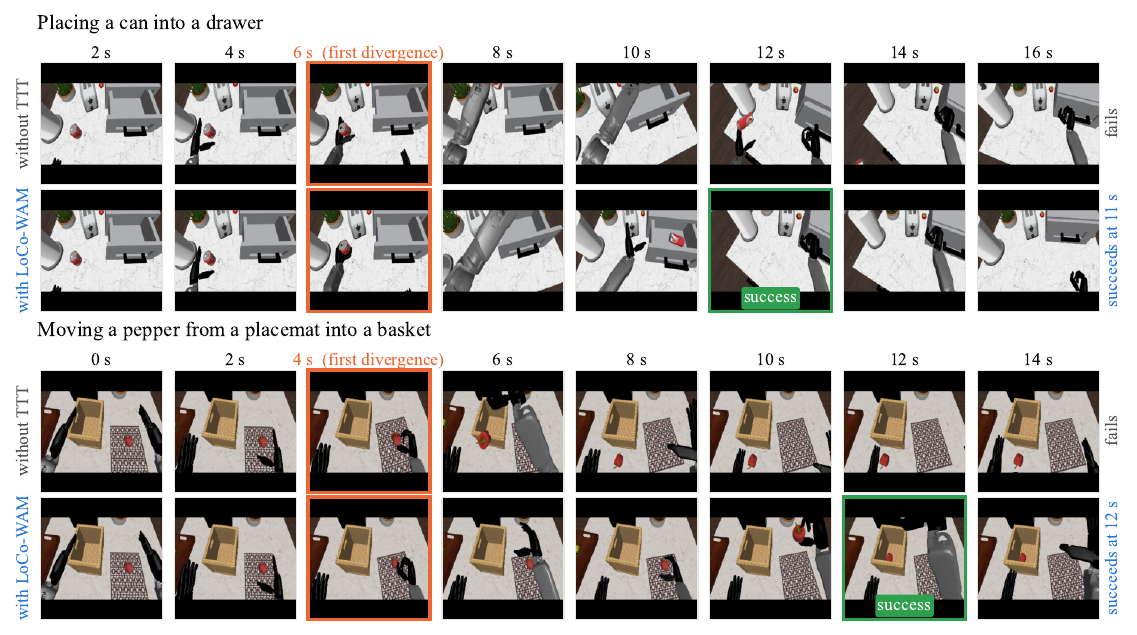}
\caption{\textbf{Episodes from the same initial state without and with LoCo-WAM.} Frames every 2\,s. The first decision at which the two runs differ and the frame at which LoCo-WAM completes the task are outlined. Long-WAM with 4.8\,s of history on RoboCasa GR-1.}
\label{fig:episodes_main}
\end{figure*}
This section analyzes how test-time training of the video expert improves the action.
In a world-action model, the update changes the prediction of the video expert in \cref{eq:loco_rollout}, and the action expert reads this prediction in \cref{eq:loco_denoise}.
We examine what the update learns from the observed frames, how it affects the action, and how this effect develops over an episode.
Unless stated otherwise, we use Long-WAM with 4.8\,s of history on RoboCasa GR-1, and the appendix reports further analysis.

\paragraph{The update improves the prediction of upcoming frames from the observed frames.}
The update of \cref{eq:update} trains the video expert on the prediction that $\operatorname{Rollout}$ performs in \cref{eq:loco_rollout}, and we first examine whether it improves this prediction on frames that it has not been trained on.
At every decision $t$, we measure the relative reduction of the video loss on the next chunk, $1 - \mathcal{L}_\text{video}(\Phi_v + \Theta_t) / \mathcal{L}_\text{video}(\Phi_v)$, before this chunk is used for the update.
In \cref{fig:mechanism}a, the reduction is close to zero at the start of an episode and increases toward its end, reaching 4.9\%, 2.5\%, 1.2\% and 1.0\% in the last tenth for 0, 2.4, 4.8 and 9.6\,s of history, which confirms such an improvement.
We then determine whether the gain originates from the observed frames or from the update itself, by replacing the targets $z$ of \cref{eq:video_loss}, the latents of these frames, with the prediction of the video expert from the same history $H$.
With these targets, the success rate is 70.8\%, the level of the base model at 71.1\%, which indicates that the gain requires the observed frames.

\begin{table}[t]
\centering
\footnotesize
\setlength{\tabcolsep}{4pt}
\begin{tabular}{@{}l c c c@{}}
\toprule
& \multicolumn{2}{c}{Video weights used in} & \\
\cmidrule(lr){2-3}
& Rollout, \cref{eq:loco_rollout} & Denoise, \cref{eq:loco_denoise} & Success \\
\midrule
Base model & $\Phi_v$ & $\Phi_v$ & 71.1\std{3.5} \\
Rollout only & $\Phi_v + \Theta_t$ & $\Phi_v$ & 77.6\std{1.7} \\
Denoise only & $\Phi_v$ & $\Phi_v + \Theta_t$ & 75.4\std{1.9} \\
LoCo-WAM & $\Phi_v + \Theta_t$ & $\Phi_v + \Theta_t$ & \textbf{78.3}\std{1.9} \\
\bottomrule
\end{tabular}
\caption{\textbf{Where the update reaches the action, RoboCasa GR-1.} The update $\Theta_t$ is trained as in LoCo-WAM and applied at inference in $\operatorname{Rollout}$, in $\operatorname{Denoise}$, or in both; the action expert $\Phi_a$ is unchanged. Long-WAM with 4.8\,s of history.}
\label{tab:readout}
\end{table}
\paragraph{The update affects the action through the imagined future and the features.}
In \cref{eq:loco_denoise}, the adapted video expert enters $\operatorname{Denoise}$ twice, as the source of the imagined future $\hat{z}_t$ and as the network that computes the features of the history and this future, which the action expert reads.
We separate the two by applying $\Theta_t$ at inference only in $\operatorname{Rollout}$ or only in $\operatorname{Denoise}$, with the update trained as in LoCo-WAM and $\Phi_a$ unchanged.
\Cref{tab:readout} reports a success rate of 77.6\% with $\Theta_t$ in $\operatorname{Rollout}$ alone and 75.4\% with $\Theta_t$ in $\operatorname{Denoise}$ alone, against 71.1\% for the base model and 78.3\% with $\Theta_t$ in both.
Both therefore contribute to the gain, the imagined future more than the features.

\paragraph{The gain depends on updating throughout the episode.}
The improvement of the prediction in \cref{fig:mechanism}a increases over an episode, and we study whether the change of the action also increases as the updates accumulate.
At every decision $t$, we measure this change as the L2 distance between the action chunks $a_t$ sampled with $\Phi_v + \Theta_t$ and with $\Phi_v$ for the same history and noise.
In \cref{fig:mechanism}b, this distance increases toward the end of the episode for every history length, whereas a random direction with the norm of $\Theta_t$, averaged over the history lengths, keeps it at its initial value, which shows that the change of the action accumulates with the learned updates.
Since such a random direction leaves the action nearly unchanged, we scale it to match the change of the imagined future under $\Theta_t$ and evaluate whether, at this size, a random change also yields the gain.
The scaled random direction reaches 71.1\%, the level of the base model, which indicates that the gain requires the change learned by the update.
Finally, we examine whether an adapter learned early in the episode suffices, by stopping the update after 4, 12 or 24 decisions and keeping $\Theta_t$ fixed afterward.
The success rate then decreases to 72.9\%, 71.9\% and 71.7\%, and the gain therefore depends on updating throughout the episode.
\Cref{fig:episodes_main} shows two episodes from the same initial state in which the trajectories separate after a few seconds and only LoCo-WAM completes the task.

%% file: tables/tab_gr1.tex
\begin{table*}[t]
\centering
\footnotesize
\setlength{\tabcolsep}{5pt}
\begin{tabular}{@{}l c c l r c c@{}}
\toprule
Method & Pick-and-place & Articulated & All & $\Delta$ & Latency (s) & Memory (GB) \\
\midrule
\multicolumn{7}{@{}l}{\textit{Prior methods}} \\
Fast-WAM (reproduced)~\cite{mogap} & 33.8 & 53.7 & 38.8 & & & \\
GR00T-N1.5~\cite{dit4dit} & 41.2 & 44.0 & 41.8 & & & \\
GR00T-N1.6~\cite{mogap} & 55.4 & 24.2 & 47.6 & & & \\
DiT4DiT~\cite{mogap} & 50.9 & 50.3 & 50.8 & & & \\
Cosmos3-Nano (reproduced)~\cite{mogap} & 48.3 & 64.3 & 52.3 & & & \\
ABot-M0~\cite{cfwam} & 57.1 & 61.7 & 58.3 & & & \\
OpenWAM-$\alpha$~\cite{openwam} & -- & -- & 60.5 & & & \\
MoGAP~\cite{mogap} & 60.8 & 66.7 & 62.3 & & & \\
JoyAI-RA~\cite{cfwam} & 60.3 & 71.7 & 63.2 & & & \\
Fast-WAM~\cite{cfwam} & 73.6 & 67.3 & 72.0 & & & \\
WALA~\cite{cfwam} & 73.4 & 80.3 & 75.2 & & & \\
CF-WAM (+EgoDex data)~\cite{cfwam} & 85.0 & 75.0 & 82.5 & & & \\
\midrule
\multicolumn{7}{@{}l}{\textit{Long-WAM, 0\,s history}} \\
\quad Base model & 68.7 & 49.4 & 63.9\std{2.5}\pub{63.3} & & 0.57 & 13.6 \\
\quad LoCo-WAM & 69.4 & 56.7 & 66.2\std{1.1} & +2.3 & 0.73 & 14.6 \\
\multicolumn{7}{@{}l}{\textit{Long-WAM, 2.4\,s history}} \\
\quad Base model & 70.9 & 58.9 & 67.9\std{0.8}\pub{66.3} & & 0.57 & 13.7 \\
\quad AdaWorldPolicy~\cite{adaworldpolicy} & 75.4 & 55.6 & 70.4\std{1.1} & +2.5 & 0.90 & 15.4 \\
\quad RoboTTT~\cite{robottt} & 76.3 & 57.8 & 71.7\std{2.7} & +3.8 & 1.06 & 14.4 \\
\quad RoboTTT w/o memory & 75.7 & 60.0 & 71.8\std{3.8} & +3.9 & 0.92 & 14.4 \\
\quad LoCo-WAM & 78.3 & 59.4 & \textbf{73.6}\std{0.5} & +5.7 & 0.76 & 14.6 \\
\quad LoCo-WAM (async) & 78.3 & 64.4 & 74.9\std{1.7} & +7.0 & 0.62 & 14.6 \\
\multicolumn{7}{@{}l}{\textit{Long-WAM, 4.8\,s history}} \\
\quad Base model & 74.8 & 60.0 & 71.1\std{3.5}\pub{71.2} & & 0.60 & 13.8 \\
\quad LoCo-WAM & 83.3 & 63.4 & \textbf{78.3}\std{1.9} & +7.2 & 0.77 & 14.5 \\
\quad LoCo-WAM (async) & 81.9 & 62.2 & 76.9\std{2.3} & +5.8 & 0.64 & 14.5 \\
\multicolumn{7}{@{}l}{\textit{Long-WAM, 9.6\,s history}} \\
\quad Base model & 79.8 & 54.5 & 73.5\std{2.7}\pub{74.5} & & 0.67 & 14.0 \\
\quad LoCo-WAM & 77.9 & 63.3 & 74.3\std{1.7} & +0.8 & 0.89 & 15.9 \\
\bottomrule
\end{tabular}
\caption{\textbf{RoboCasa GR-1.} On 18 pick-and-place tasks and 6 articulated tasks that place an object into a drawer, cabinet or microwave and close it. Latency: time from observation to action per decision on one GPU, including the LoRA update, which the asynchronous variant runs during the execution of the previous action chunk. Memory: peak GPU memory.}
\label{tab:gr1}
\end{table*}

%% file: tables/tab_rc365.tex
\begin{table}[t]
\centering
\footnotesize
\setlength{\tabcolsep}{3pt}
\resizebox{\columnwidth}{!}{%
\begin{tabular}{@{}l c c c l r@{}}
\toprule
Method & Atomic & \begin{tabular}[b]{@{}c@{}}Seen\\composite\end{tabular} & \begin{tabular}[b]{@{}c@{}}Unseen\\composite\end{tabular} & All & $\Delta$ \\
\midrule
\multicolumn{6}{@{}l}{\textit{Prior methods}} \\
Diffusion Policy~\cite{diffusionpolicy} & 15.7 & 0.2 & 1.3 & 6.1 & \\
$\pi_0$~\cite{pi0} & 34.6 & 6.1 & 1.1 & 14.8 & \\
$\pi_{0.5}$~\cite{pi05} & 39.6 & 7.1 & 1.2 & 16.9 & \\
GR00T-N1.6 & 51.1 & 9.4 & 1.7 & 21.9 & \\
GR00T-N1.5 & 50.7 & 14.8 & 2.7 & 23.9 & \\
Qwen-RobotManip~\cite{longwam} & 68.6 & 20.1 & 14.9 & 35.9 & \\
ABot-M0.5 & 75.6 & 37.7 & 3.3 & 40.3 & \\
PatchWAM~\cite{patchwam} & 78.7 & 46.4 & 13.8 & 47.6 & \\
\midrule
\multicolumn{6}{@{}l}{\textit{Long-WAM, 2.4\,s history}} \\
\quad Base model & 68.0 & 14.9 & 5.1 & 30.9\std{0.0}\pub{31.4} & \\
\quad AdaWorldPolicy & 67.8 & 15.2 & 5.2 & 30.9\std{0.1} & +0.0 \\
\quad LoCo-WAM & 69.2 & 16.2 & 6.7 & \textbf{32.2}\std{0.2} & +1.4 \\
\bottomrule
\end{tabular}}
\caption{\textbf{RoboCasa365.} On 18 atomic tasks and 16 composite tasks seen during baseline model training, and 16 composite tasks not seen during baseline training.}
\label{tab:rc365}
\end{table}

%% file: tables/tab_liberomem.tex
\begin{table}[t]
\centering
\footnotesize
\setlength{\tabcolsep}{5pt}
\begin{tabular}{@{}lcccc@{}}
\toprule
Method & Success & $\Delta$ & Subgoal & $\Delta$ \\
\midrule
\multicolumn{5}{@{}l}{\textit{Prior methods}} \\
$\pi_0$~\cite{twoam} & 12.3 & & 70.8 & \\
2AM~\cite{twoam} & 11.8 & & 76.3 & \\
\midrule
\multicolumn{5}{@{}l}{\textit{LingBot-VA, 30-chunk history}} \\
\quad Base model & 32.2\std{1.8} & & 85.8\std{2.1} & \\
\multicolumn{5}{@{}l}{\textit{LingBot-VA, 8-chunk history}} \\
\quad Base model & 27.5\std{2.0} & & 82.7\std{1.2} & \\
\quad AdaWorldPolicy~\cite{adaworldpolicy} & 27.7\std{3.3} & +0.2 & 82.9\std{0.8} & +0.2 \\
\quad ReCAP~\cite{recap} & 24.7\std{1.5} & $-$2.8 & 84.8\std{0.6} & +2.1 \\
\quad RoboTTT~\cite{robottt} & 27.8\std{1.0} & +0.3 & 83.1\std{2.3} & +0.4 \\
\quad RoboTTT w/o memory & 29.2\std{1.0} & +1.7 & 84.5\std{1.2} & +1.7 \\
\quad Future Consensus~\cite{futureconsensus} & 30.3\std{3.3} & +2.8 & 85.5\std{2.0} & +2.8 \\
\quad LoCo-WAM & \textbf{32.2}\std{0.3} & +4.7 & \textbf{85.8}\std{1.0} & +3.1 \\
\bottomrule
\end{tabular}
\caption{\textbf{LIBERO-Mem.} Success: all subgoals completed without overshooting. Subgoal: subgoal completion rate.}
\label{tab:liberomem}
\end{table}

%% file: tables/tab_robotwin.tex
\begin{table}[t]
\centering
\footnotesize
\setlength{\tabcolsep}{3pt}
\begin{tabular}{@{}l c c c@{}}
\toprule
Method & Long-context (4) & Other (4) & All (8) \\
\midrule
\multicolumn{4}{@{}l}{\textit{Prior methods}} \\
$\pi_0$, from~\cite{lingbotva} & 25.8 & 59.5 & 42.6 \\
$\pi_{0.5}$, from~\cite{lingbotva} & 50.3 & 69.8 & 60.0 \\
X-VLA~\cite{xvla}, from~\cite{motus} & 58.5 & 69.3 & 63.9 \\
Fast-WAM~\cite{fastwam} & 67.8 & 80.8 & 74.3 \\
AHA-WAM~\cite{ahawam} & 73.3 & 82.8 & 78.0 \\
Motus~\cite{motus} & 77.5 & 81.0 & 79.3 \\
HiLo-WAM-k8~\cite{hilowam} & 80.5 & 79.0 & 79.8 \\
GeoMem-VLA~\cite{geomemvla} & 76.5 & 85.0 & 80.8 \\
Rolling-WAM~\cite{rollingwam} & 79.5 & 84.5 & 82.0 \\
StageWAM~\cite{stagewam} & 79.5 & 86.3 & 82.9 \\
LingBot-VA (reported)~\cite{lingbotva} & 62.8 & 86.3 & 74.5 \\
LingBot-VA (re-eval.)~\cite{lawam} & 53.3 & 78.3 & 65.8 \\
\midrule
\multicolumn{4}{@{}l}{\textit{LingBot-VA, 72-chunk history (default)}} \\
\quad Base model & 54.2 & 77.7 & 66.0\std{0.8} \\
\quad AdaWorldPolicy & 54.1 & 78.3 & 66.2\std{0.6} \\
\quad LoCo-WAM & 55.1 & 79.1 & \textbf{67.1}\std{3.0} \\
\multicolumn{4}{@{}l}{\textit{LingBot-VA, 16-chunk history}} \\
\quad Base model & 51.8 & 77.3 & 64.5\std{1.2} \\
\quad LoCo-WAM & 53.3 & 77.3 & \textbf{65.3}\std{1.7} \\
\bottomrule
\end{tabular}
\caption{\textbf{RoboTwin 2.0.} On the 8 hardest tasks: the 4 tasks on which a 72-chunk history outperforms a 16-chunk history, the other 4 tasks, and all 8.}
\label{tab:robotwin}
\end{table}

%% file: tables/tab_ablation_main.tex
\begin{table}[t]
\centering
\footnotesize
\setlength{\tabcolsep}{4pt}
\begin{tabular}{@{}l l@{}}
\toprule
Ablation & Success \\
\midrule
LoRA rank $r$ (8 / 16 / 64) & 73.2 / 73.4 / 74.1 \\
Updates per decision $S_v$ (2 / 4) & 73.6 / 73.9 \\
Learning rate $\eta_v$ ($\times 1/3$ / $\times 5/3$) & 72.8 / 77.7 \\
LoRA placement (SA only / FFN only) & 71.4 / 72.8 \\
\bottomrule
\end{tabular}
\caption{\textbf{Ablations on RoboCasa GR-1.} Long-WAM with 4.8\,s of history.}
\label{tab:ablation_main}
\end{table}

%% file: tables/tab_side.tex
\begin{table}[t]
\centering
\footnotesize
\setlength{\tabcolsep}{4pt}
\begin{tabular}{@{}l c c@{}}
\toprule
Loss of the update & 2.4\,s & 4.8\,s \\
\midrule
None, base model & 67.9\std{0.8} & 71.1\std{3.5} \\
$\mathcal{L}_{\text{action}}$, \cref{eq:action_loss} & 72.2\std{2.1} & 73.3\std{0.4} \\
$\mathcal{L}_{\text{video}} + \mathcal{L}_{\text{action}}$, \cref{eq:video_loss} + \cref{eq:action_loss} & 72.6\std{1.5} & 73.7\std{2.2} \\
$\mathcal{L}_{\text{video}}$, \cref{eq:video_loss}, LoCo-WAM & \textbf{73.6}\std{0.5} & \textbf{78.3}\std{1.9} \\
\bottomrule
\end{tabular}
\caption{\textbf{Which loss to update with, RoboCasa GR-1.} Each loss trains a low-rank adapter on the expert it supervises, the video expert for $\mathcal{L}_{\text{video}}$ and the action expert for $\mathcal{L}_{\text{action}}$. Long-WAM with 2.4 and 4.8\,s of history.}
\label{tab:side}
\end{table}

%% file: sec/5_conclusion.tex
\section{Conclusion}
\label{sec:conclusion}
We presented LoCo-WAM, which trains the video expert of a world-action model at test time with its own video loss on the observed frames, while keeping the architecture and the history length of the model.
On two world-action models and four simulation benchmarks, LoCo-WAM improves the base model on all of them, and on RoboCasa GR-1, a model with 2.4\,s of history matches the base model with 9.6\,s, and a model with 4.8\,s exceeds it by 4.8 points.
Our analysis shows that the update learns from the observed frames to predict upcoming frames, and that it affects the action mainly through the imagined future.
This change of the action accumulates over the episode and depends on updating throughout it.
In a world-action model, the environment provides the targets of the video loss at every decision, and LoCo-WAM shows that training on these targets at test time enables a short history to reach the performance of a longer one.

%% file: sec/X_suppl.tex
\clearpage
\onecolumn
\appendix
\begin{center}
{\Large\bfseries Appendix}
\end{center}

\section{Additional results}
\label{sec:appx_results}

\begin{figure}[!htbp]
\centering
\includegraphics[width=\textwidth]{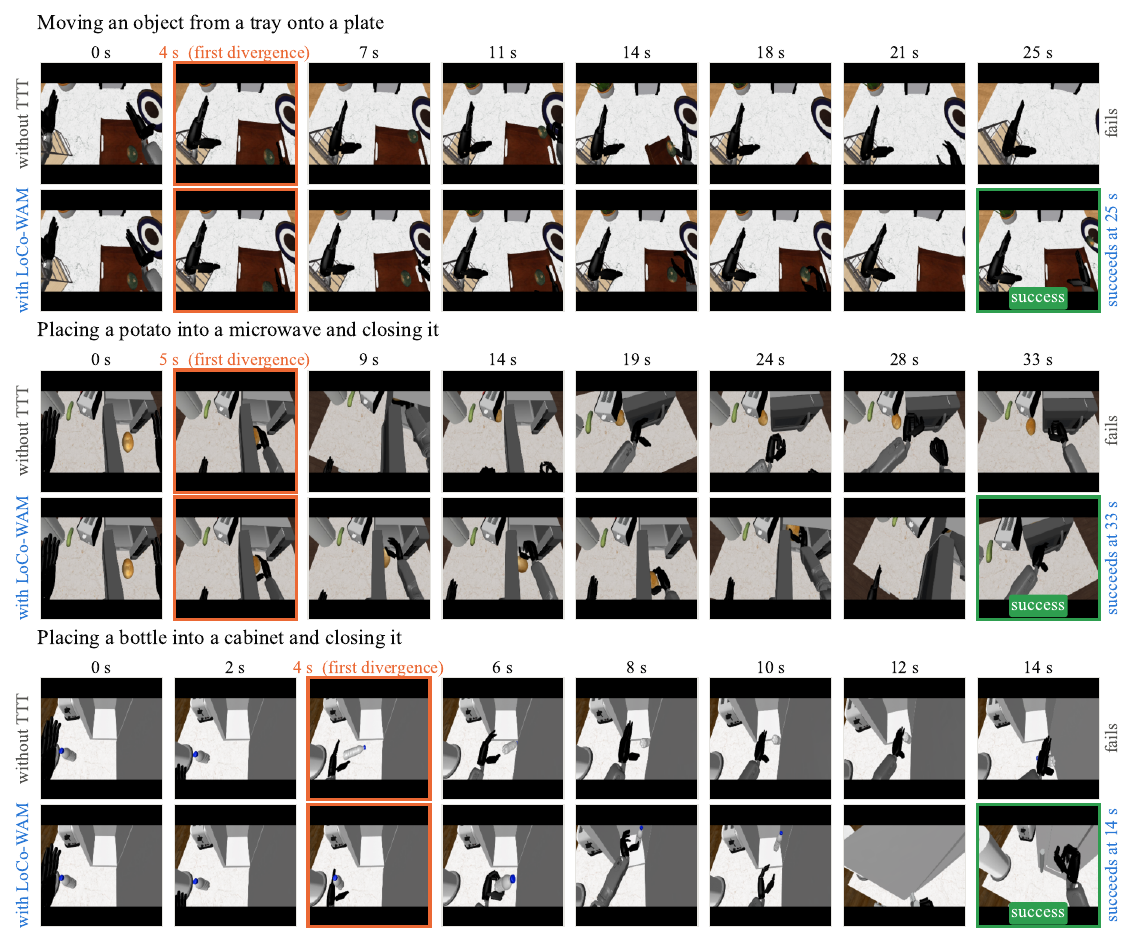}
\caption{\textbf{More episodes from the same initial state without and with LoCo-WAM.} The first decision at which the two runs differ and the frame at which LoCo-WAM completes the task are outlined, and the frames are spaced evenly up to this frame. Long-WAM with 4.8\,s of history on RoboCasa GR-1.}
\label{fig:episodes_appx}
\end{figure}

\begin{figure}[!htbp]
\centering
\includegraphics[width=\textwidth]{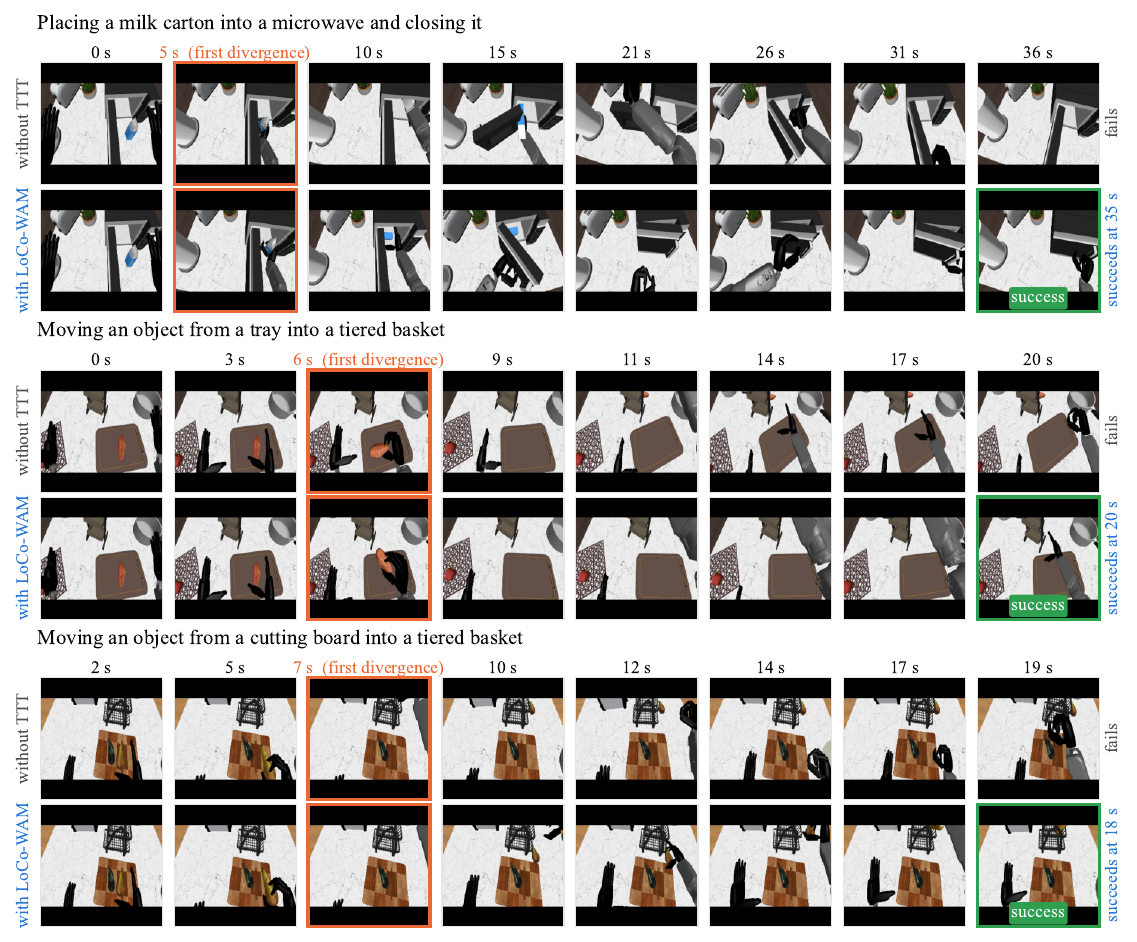}
\caption{\textbf{More episodes from the same initial state without and with LoCo-WAM, continued.} Same layout as \cref{fig:episodes_appx}.}
\label{fig:episodes_appx_b}
\end{figure}

\paragraph{More visualization results.}
\Cref{fig:episodes_appx,fig:episodes_appx_b} show six further pairs of episodes from the same initial state without and with LoCo-WAM.

\FloatBarrier
\section{Additional analysis}
\label{sec:appx_analysis}

\paragraph{Resetting the adapter at every episode.}
LoCo-WAM resets the adapter at the start of every episode, so that each episode is adapted only to its own observed frames.
A natural alternative keeps the adapter across the episodes of a task, as AdaWorldPolicy does with its world model.
On LingBot-VA with LIBERO-Mem, this alternative lowers the success rate by 3.7 points and the subgoal completion rate by 4.0 points relative to the base model in the same runs, and all three runs fall below the base model.
The adaptation to one episode therefore does not carry over to later episodes, which is why LoCo-WAM resets the adapter.

\paragraph{A larger step does not help the action update.}
In \cref{tab:side}, updating the action expert alone improves the base model less than updating the video expert, and it also takes much smaller steps, changing the adapter by 0.002 in norm per update compared with 0.059 and 0.035 for the video update with 2.4 and 4.8\,s of history.
When we scale its learning rate to match these steps, the action update reaches 63.2\% and 69.3\% on RoboCasa GR-1, against 71.0\% and 69.6\% for the base model in the same runs, and thus no longer improves the base model.
The smaller step therefore does not account for the lower success rate of the action update, which supports updating the video expert in LoCo-WAM.

%% file: main.bbl
\begin{thebibliography}{62}
\providecommand{\natexlab}[1]{#1}
\providecommand{\url}[1]{\texttt{#1}}
\expandafter\ifx\csname urlstyle\endcsname\relax
  \providecommand{\doi}[1]{doi: #1}\else
  \providecommand{\doi}{doi: \begingroup \urlstyle{rm}\Url}\fi

\bibitem[Bai et~al.(2025)Bai, Gao, and Shou]{evolvevla}
Zechen Bai, Chen Gao, and Mike~Zheng Shou.
\newblock Evolve-vla: Test-time training from environment feedback for vision-language-action models.
\newblock \emph{arXiv preprint arXiv:2512.14666}, 2025.

\bibitem[Behrouz et~al.(2024)Behrouz, Zhong, and Mirrokni]{titans}
Ali Behrouz, Peilin Zhong, and Vahab Mirrokni.
\newblock Titans: Learning to memorize at test time.
\newblock \emph{arXiv preprint arXiv:2501.00663}, 2024.

\bibitem[Bi et~al.(2025)Bi, Tan, Xie, Wang, Huang, Liu, Zhao, Feng, Xiang, Rong, et~al.]{motus}
Hongzhe Bi, Hengkai Tan, Shenghao Xie, Zeyuan Wang, Shuhe Huang, Haitian Liu, Ruowen Zhao, Yao Feng, Chendong Xiang, Yinze Rong, et~al.
\newblock Motus: A unified latent action world model.
\newblock \emph{arXiv preprint arXiv:2512.13030}, 2025.

\bibitem[Bjorck et~al.(2025)Bjorck, Casta{\~n}eda, Cherniadev, Da, Ding, Fan, Fang, Fox, Hu, Huang, et~al.]{gr00tn1}
Johan Bjorck, Fernando Casta{\~n}eda, Nikita Cherniadev, Xingye Da, Runyu Ding, Linxi Fan, Yu Fang, Dieter Fox, Fengyuan Hu, Spencer Huang, et~al.
\newblock Gr00t n1: An open foundation model for generalist humanoid robots.
\newblock \emph{arXiv preprint arXiv:2503.14734}, 2025.

\bibitem[Black et~al.(2024)Black, Brown, Driess, Esmail, Equi, Finn, Fusai, Groom, Hausman, Ichter, et~al.]{pi0}
Kevin Black, Noah Brown, Danny Driess, Adnan Esmail, Michael Equi, Chelsea Finn, Niccolo Fusai, Lachy Groom, Karol Hausman, Brian Ichter, et~al.
\newblock $\pi_0$: A vision-language-action flow model for general robot control.
\newblock \emph{arXiv preprint arXiv:2410.24164}, 2024.

\bibitem[Cai et~al.(2026)Cai, Ling, Chu, Liu, Kang, Liang, Xu, Mao, Zhang, Yang, et~al.]{ahawam}
Jisong Cai, Long Ling, Shiwei Chu, Zhongshan Liu, Jiayue Kang, Zhixuan Liang, Wenjie Xu, Yinan Mao, Weinan Zhang, Xiaokang Yang, et~al.
\newblock Aha-wam:asynchronous horizon-adaptive world-action modeling with observation-guided context routing.
\newblock \emph{arXiv preprint arXiv:2606.09811}, 2026.

\bibitem[Chen et~al.(2026)Chen, Wang, Chen, Chen, Gao, Tang, Li, Liu, Yao, Li, et~al.]{lawam}
Jialei Chen, Kai Wang, Kang Chen, Shuaihang Chen, Feng Gao, Wenhao Tang, Zhiyuan Li, Weilin Liu, Zhuyu Yao, Boxun Li, et~al.
\newblock Lawam: Latent world action models for efficient dynamics-aware robot policies.
\newblock \emph{arXiv preprint arXiv:2606.15768}, 2026.

\bibitem[Chen et~al.(2025)Chen, Chen, Chen, Cai, Liu, Li, Liang, Lin, Ge, Gu, et~al.]{robotwin2}
Tianxing Chen, Zanxin Chen, Baijun Chen, Zijian Cai, Yibin Liu, Zixuan Li, Qiwei Liang, Xianliang Lin, Yiheng Ge, Zhenyu Gu, et~al.
\newblock Robotwin 2.0: A scalable data generator and benchmark with strong domain randomization for robust bimanual robotic manipulation.
\newblock \emph{arXiv preprint arXiv:2506.18088}, 2025.

\bibitem[Cheng et~al.(2025)Cheng, Kong, Xie, Lee, Wu, and Yang]{cheng2025latent}
Sheng Cheng, Deqian Kong, Jianwen Xie, Kookjin Lee, Ying~Nian Wu, and Yezhou Yang.
\newblock Latent space energy-based neural {ODEs}.
\newblock \emph{Transactions on Machine Learning Research}, 2025.

\bibitem[Chi et~al.(2023)Chi, Xu, Feng, Cousineau, Du, Burchfiel, Tedrake, and Song]{diffusionpolicy}
Cheng Chi, Zhenjia Xu, Siyuan Feng, Eric Cousineau, Yilun Du, Benjamin Burchfiel, Russ Tedrake, and Shuran Song.
\newblock Diffusion policy: Visuomotor policy learning via action diffusion.
\newblock \emph{arXiv preprint arXiv:2303.04137}, 2023.

\bibitem[Chung et~al.(2025)Chung, Hanyu, Nguyen, Le, Bumgarner, Nguyen, Vo, Yamazaki, Rainwater, Kieu, et~al.]{liberomem}
Nhat Chung, Taisei Hanyu, Toan Nguyen, Huy Le, Frederick Bumgarner, Duy Minh~Ho Nguyen, Khoa Vo, Kashu Yamazaki, Chase Rainwater, Tung Kieu, et~al.
\newblock Rethinking progression of memory state in robotic manipulation: An object-centric perspective.
\newblock \emph{arXiv preprint arXiv:2511.11478}, 2025.

\bibitem[Dalal et~al.(2025)Dalal, Koceja, Hussein, Xu, Zhao, Song, Han, Cheung, Kautz, Guestrin, et~al.]{oneminute}
Karan Dalal, Daniel Koceja, Gashon Hussein, Jiarui Xu, Yue Zhao, Youjin Song, Shihao Han, Ka~Chun Cheung, Jan Kautz, Carlos Guestrin, et~al.
\newblock One-minute video generation with test-time training.
\newblock \emph{arXiv preprint arXiv:2504.05298}, 2025.

\bibitem[Du et~al.(2023)Du, Yang, Dai, Dai, Nachum, Tenenbaum, Schuurmans, and Abbeel]{unipi}
Yilun Du, Mengjiao Yang, Bo Dai, Hanjun Dai, Ofir Nachum, Joshua~B. Tenenbaum, Dale Schuurmans, and Pieter Abbeel.
\newblock Learning universal policies via text-guided video generation.
\newblock \emph{arXiv preprint arXiv:2302.00111}, 2023.

\bibitem[Feng et~al.(2026)Feng, Han, Lyu, Liu, Zheng, Wan, Liu, Han, Li, Zhang, et~al.]{wamttt}
Yusen Feng, Bingchen Han, Jiangran Lyu, Kai Liu, Yixin Zheng, Yuxuan Wan, Weiheng Liu, Sun Han, Ruiqin Li, Yulong Zhang, et~al.
\newblock Wam-ttt: Steering world-action models by watching human play at test time.
\newblock \emph{arXiv preprint arXiv:2607.06988}, 2026.

\bibitem[Gandelsman et~al.(2022)Gandelsman, Sun, Chen, and Efros]{tttmae}
Yossi Gandelsman, Yu Sun, Xinlei Chen, and Alexei~A. Efros.
\newblock Test-time training with masked autoencoders.
\newblock \emph{arXiv preprint arXiv:2209.07522}, 2022.

\bibitem[Hansen et~al.(2020)Hansen, Jangir, Sun, Alenyà, Abbeel, Efros, Pinto, and Wang]{pad}
Nicklas Hansen, Rishabh Jangir, Yu Sun, Guillem Alenyà, Pieter Abbeel, Alexei~A. Efros, Lerrel Pinto, and Xiaolong Wang.
\newblock Self-supervised policy adaptation during deployment.
\newblock \emph{arXiv preprint arXiv:2007.04309}, 2020.

\bibitem[Hong et~al.(2024)Hong, Liu, Wu, Zhai, Chang, Li, Lin, Lin, Wang, Yang, et~al.]{slowfastvgen}
Yining Hong, Beide Liu, Maxine Wu, Yuanhao Zhai, Kai-Wei Chang, Linjie Li, Kevin Lin, Chung-Ching Lin, Jianfeng Wang, Zhengyuan Yang, et~al.
\newblock Slowfast-vgen: Slow-fast learning for action-driven long video generation.
\newblock \emph{arXiv preprint arXiv:2410.23277}, 2024.

\bibitem[Hu et~al.(2022)Hu, Shen, Wallis, Allen-Zhu, Li, Wang, Wang, and Chen]{lora}
Edward~J. Hu, Yelong Shen, Phillip Wallis, Zeyuan Allen-Zhu, Yuanzhi Li, Shean Wang, Lu Wang, and Weizhu Chen.
\newblock {LoRA}: Low-rank adaptation of large language models.
\newblock In \emph{International Conference on Learning Representations}, 2022.

\bibitem[Hu et~al.(2024)Hu, Guo, Wang, Chen, Wang, Zhang, Sreenath, Lu, and Chen]{vpp}
Yucheng Hu, Yanjiang Guo, Pengchao Wang, Xiaoyu Chen, Yen-Jen Wang, Jianke Zhang, Koushil Sreenath, Chaochao Lu, and Jianyu Chen.
\newblock Video prediction policy: A generalist robot policy with predictive visual representations.
\newblock \emph{arXiv preprint arXiv:2412.14803}, 2024.

\bibitem[Hu et~al.(2026)Hu, Chen, Cao, and Detry]{twoam}
Yutong Hu, Fengjiao Chen, Xuezhi Cao, and Renaud Detry.
\newblock 2am: Grounding agent-side memory as guidance for steerable action models in long-horizon manipulation.
\newblock \emph{arXiv preprint arXiv:2609.11308}, 2026.

\bibitem[Huang et~al.(2026)Huang, Zhang, Liu, Liu, Yang, Mao, Wang, Xiao, Lin, Hu, et~al.]{longwam}
Wei Huang, Bohan Zhang, Chenzhi Liu, Isabella Liu, Shuai Yang, Weian Mao, Luozhou Wang, Yicheng Xiao, Weifeng Lin, Qixin Hu, et~al.
\newblock Long-wam: Scaling the context of world-action models.
\newblock \emph{arXiv preprint arXiv:2610.10528}, 2026.

\bibitem[Intelligence et~al.(2025)Intelligence, Black, Brown, Darpinian, Dhabalia, Driess, Esmail, Equi, Finn, Fusai, et~al.]{pi05}
Physical Intelligence, Kevin Black, Noah Brown, James Darpinian, Karan Dhabalia, Danny Driess, Adnan Esmail, Michael Equi, Chelsea Finn, Niccolo Fusai, et~al.
\newblock $\pi_{0.5}$: a vision-language-action model with open-world generalization.
\newblock \emph{arXiv preprint arXiv:2504.16054}, 2025.

\bibitem[Jiang et~al.(2026)Jiang, Chebotar, Zheng, Hu, Ge, Wu, Dai, Reed, Fei-Fei, Zhu, et~al.]{robottt}
Yunfan Jiang, Yevgen Chebotar, Ruijie Zheng, Fengyuan Hu, Yunhao Ge, Jimmy Wu, Tianyuan Dai, Scott Reed, Li Fei-Fei, Yuke Zhu, et~al.
\newblock Robottt: Context scaling for robot policies.
\newblock \emph{arXiv preprint arXiv:2607.15275}, 2026.

\bibitem[Kim et~al.(2026{\natexlab{a}})Kim, Gao, Lin, Lin, Ge, Lam, Liang, Song, Liu, Finn, et~al.]{cosmospolicy}
Moo~Jin Kim, Yihuai Gao, Tsung-Yi Lin, Yen-Chen Lin, Yunhao Ge, Grace Lam, Percy Liang, Shuran Song, Ming-Yu Liu, Chelsea Finn, et~al.
\newblock Cosmos policy: Fine-tuning video models for visuomotor control and planning.
\newblock \emph{arXiv preprint arXiv:2601.16163}, 2026{\natexlab{a}}.

\bibitem[Kim et~al.(2026{\natexlab{b}})Kim, Sung, Kwon, Kim, and Shin]{mogap}
Taeyoung Kim, Junyoung Sung, Heeseung Kwon, Minseop Kim, and Jinwoo Shin.
\newblock Adapting video models for robot control via motion-guided asymmetric prediction.
\newblock \emph{arXiv preprint arXiv:2610.36624}, 2026{\natexlab{b}}.

\bibitem[Kong et~al.(2024)Kong, Xu, Zhao, Pang, Xie, Lizarraga, Huang, Xie, and Wu]{kong2024latent}
Deqian Kong, Dehong Xu, Minglu Zhao, Bo Pang, Jianwen Xie, Andrew Lizarraga, Yuhao Huang, Sirui Xie, and Ying~Nian Wu.
\newblock Latent plan transformer for trajectory abstraction: Planning as latent space inference.
\newblock \emph{Advances in Neural Information Processing Systems}, 37:\penalty0 123379--123401, 2024.

\bibitem[Kong et~al.(2026)Kong, Sun, Cheng, Xie, Pang, Xie, Geng, Ding, and Wu]{kong2026plan}
Deqian Kong, Guangyan Sun, Sheng Cheng, Sirui Xie, Bo Pang, Jianwen Xie, Tony Geng, Caiwen Ding, and Ying~Nian Wu.
\newblock Learning to plan from random exploration.
\newblock \emph{arXiv preprint arXiv:2609.38383}, 2026.

\bibitem[Krause et~al.(2018)Krause, Kahembwe, Murray, and Renals]{dynamiceval}
Ben Krause, Emmanuel Kahembwe, Iain Murray, and Steve Renals.
\newblock Dynamic evaluation of neural sequence models.
\newblock In \emph{Proceedings of the 35th International Conference on Machine Learning}, pages 2766--2775, 2018.

\bibitem[Lee et~al.(2026)Lee, Seo, Jang, and Gong]{salt}
Ahin Lee, Jinwoo Seo, Youngsoo Jang, and Taesik Gong.
\newblock Adapting vision-language-action models to unknown visual disruptions during execution.
\newblock \emph{arXiv preprint arXiv:2610.07946}, 2026.

\bibitem[Li et~al.(2026)Li, Zhang, Luo, Yang, Wang, Han, Yu, Gao, Xue, Zhu, et~al.]{lingbotva}
Lin Li, Qihang Zhang, Yiming Luo, Shuai Yang, Ruilin Wang, Fei Han, Mingrui Yu, Zelin Gao, Nan Xue, Xing Zhu, et~al.
\newblock Causal world modeling for robot control.
\newblock \emph{arXiv preprint arXiv:2601.21998}, 2026.

\bibitem[Li et~al.(2025)Li, Gao, Sadigh, and Song]{uva}
Shuang Li, Yihuai Gao, Dorsa Sadigh, and Shuran Song.
\newblock Unified video action model.
\newblock \emph{arXiv preprint arXiv:2503.00200}, 2025.

\bibitem[Liu et~al.(2026{\natexlab{a}})Liu, Liu, Wang, Zhuang, Liang, Yang, Xu, Wang, Liu, and Han]{ttvla}
Changyu Liu, Yiyang Liu, Taowen Wang, Qiao Zhuang, James~Chenhao Liang, Wenhao Yang, Renjing Xu, Qifan Wang, Dongfang Liu, and Cheng Han.
\newblock On-the-fly vla adaptation via test-time reinforcement learning.
\newblock \emph{arXiv preprint arXiv:2601.06748}, 2026{\natexlab{a}}.

\bibitem[Liu et~al.(2026{\natexlab{b}})Liu, Yang, Wang, Jiang, Chi, Xu, Ding, Chen, and Wang]{stagewam}
Xiao Liu, Yuguang Yang, Xi Wang, Kai Jiang, Cheng Chi, Yong Xu, Wenchao Ding, Yilun Chen, and Yan Wang.
\newblock Stagewam: Joint-embedding stage prediction for world-action models in robot manipulation.
\newblock \emph{arXiv preprint arXiv:2608.10780}, 2026{\natexlab{b}}.

\bibitem[Ma et~al.(2026)Ma, Zheng, Wang, Jiang, Cui, Liang, and Yang]{dit4dit}
Teli Ma, Jia Zheng, Zifan Wang, Chunli Jiang, Andy Cui, Junwei Liang, and Shuo Yang.
\newblock Dit4dit: Jointly modeling video dynamics and actions for generalizable robot control.
\newblock \emph{arXiv preprint arXiv:2603.10448}, 2026.

\bibitem[McClelland et~al.(1995)McClelland, McNaughton, and O'Reilly]{mcclelland1995complementary}
James~L. McClelland, Bruce~L. McNaughton, and Randall~C. O'Reilly.
\newblock Why there are complementary learning systems in the hippocampus and neocortex: Insights from the successes and failures of connectionist models of learning and memory.
\newblock \emph{Psychological Review}, 102\penalty0 (3):\penalty0 419--457, 1995.

\bibitem[Nasiriany et~al.(2026)Nasiriany, Nasiriany, Maddukuri, and Zhu]{robocasa365}
Soroush Nasiriany, Sepehr Nasiriany, Abhiram Maddukuri, and Yuke Zhu.
\newblock Robocasa365: A large-scale simulation framework for training and benchmarking generalist robots.
\newblock \emph{arXiv preprint arXiv:2603.04356}, 2026.

\bibitem[Noh et~al.(2025)Noh, Kong, Zhao, Lizarraga, Xie, Wu, and Hong]{noh2025latent}
Donghun Noh, Deqian Kong, Minglu Zhao, Andrew Lizarraga, Jianwen Xie, Ying~Nian Wu, and Dennis Hong.
\newblock Latent adaptive planner for dynamic manipulation.
\newblock In \emph{Conference on Robot Learning}, pages 2430--2448. PMLR, 2025.

\bibitem[Park et~al.(2026)Park, Kim, In, Kim, Kang, and Park]{t3vf}
Sangwu Park, Wonjoong Kim, Yeonjun In, Sein Kim, Hongseok Kang, and Chanyoung Park.
\newblock Test-time training for visual foresight vision-language-action models.
\newblock \emph{arXiv preprint arXiv:2605.08215}, 2026.

\bibitem[Ruan et~al.(2026)Ruan, Hsiao, Lo, and Shuai]{futureconsensus}
Bo-Kai Ruan, Teng-Fang Hsiao, Ling Lo, and Hong-Han Shuai.
\newblock Is the future compatible? diagnosing dynamic consistency in world action models.
\newblock \emph{arXiv preprint arXiv:2605.07514}, 2026.

\bibitem[Sun et~al.(2019)Sun, Wang, Liu, Miller, Efros, and Hardt]{ttt_sun}
Yu Sun, Xiaolong Wang, Zhuang Liu, John Miller, Alexei~A. Efros, and Moritz Hardt.
\newblock Test-time training with self-supervision for generalization under distribution shifts.
\newblock \emph{arXiv preprint arXiv:1909.13231}, 2019.

\bibitem[Sun et~al.(2024)Sun, Li, Dalal, Xu, Vikram, Zhang, Dubois, Chen, Wang, Koyejo, et~al.]{tttlayers}
Yu Sun, Xinhao Li, Karan Dalal, Jiarui Xu, Arjun Vikram, Genghan Zhang, Yann Dubois, Xinlei Chen, Xiaolong Wang, Sanmi Koyejo, et~al.
\newblock Learning to (learn at test time): Rnns with expressive hidden states.
\newblock \emph{arXiv preprint arXiv:2407.04620}, 2024.

\bibitem[Tandon et~al.(2025)Tandon, Dalal, Li, Koceja, Rød, Buchanan, Wang, Leskovec, Koyejo, Hashimoto, et~al.]{ttte2e}
Arnuv Tandon, Karan Dalal, Xinhao Li, Daniel Koceja, Marcel Rød, Sam Buchanan, Xiaolong Wang, Jure Leskovec, Sanmi Koyejo, Tatsunori Hashimoto, et~al.
\newblock End-to-end test-time training for long context.
\newblock \emph{arXiv preprint arXiv:2512.23675}, 2025.

\bibitem[Tseng et~al.(2007)Tseng, Diedrichsen, Krakauer, Shadmehr, and Bastian]{tseng2007sensory}
Ya-weng Tseng, J{\"o}rn Diedrichsen, John~W. Krakauer, Reza Shadmehr, and Amy~J. Bastian.
\newblock Sensory prediction errors drive cerebellum-dependent adaptation of reaching.
\newblock \emph{Journal of Neurophysiology}, 98\penalty0 (1):\penalty0 54--62, 2007.

\bibitem[Wan et~al.(2026)Wan, Wang, Si, Zhang, and Li]{worldagen}
Chi Wan, Kangrui Wang, Yuan Si, Pingyue Zhang, and Manling Li.
\newblock Worldagen: Unified state-action prediction with test-time world model training.
\newblock \emph{arXiv preprint arXiv:2609.08162}, 2026.

\bibitem[Wang et~al.(2020)Wang, Shelhamer, Liu, Olshausen, and Darrell]{tent}
Dequan Wang, Evan Shelhamer, Shaoteng Liu, Bruno Olshausen, and Trevor Darrell.
\newblock Tent: Fully test-time adaptation by entropy minimization.
\newblock \emph{arXiv preprint arXiv:2006.10726}, 2020.

\bibitem[Wang et~al.(2026{\natexlab{a}})Wang, Gu, Chen, Xu, Ma, Yang, Li, Huang, Wang, and Su]{dimwam}
Kai Wang, Zhaopeng Gu, Yixiang Chen, Yuan Xu, Qisen Ma, Jiabing Yang, Zhaowen Li, Yan Huang, Liang Wang, and Peng Su.
\newblock {DIM-WAM}: World-action modeling with diverse historical event memory.
\newblock \emph{arXiv preprint arXiv:2606.27677}, 2026{\natexlab{a}}.

\bibitem[Wang et~al.(2023)Wang, Sun, Tandon, Gandelsman, Chen, Efros, and Wang]{tttvideo}
Renhao Wang, Yu Sun, Arnuv Tandon, Yossi Gandelsman, Xinlei Chen, Alexei~A. Efros, and Xiaolong Wang.
\newblock Test-time training on video streams.
\newblock \emph{arXiv preprint arXiv:2307.05014}, 2023.

\bibitem[Wang et~al.(2026{\natexlab{b}})Wang, Xie, Jia, Xu, Zhang, and Yu]{patchwam}
Tianheng Wang, Zhou Xie, Heng Jia, Jianhua Xu, Tong Zhang, and Kaicheng Yu.
\newblock An action is worth one patch: Unified world-action modeling with patchwam.
\newblock \emph{arXiv preprint arXiv:2609.25961}, 2026{\natexlab{b}}.

\bibitem[Wang et~al.(2026{\natexlab{c}})Wang, Guo, Ding, Fang, Lai, Dong, Wan, Chen, Gao, and Xu]{hilowam}
Xingyu Wang, Yanjiang Guo, Zhijing Ding, Jiaxin Fang, Zeyu Lai, Jiayu Dong, Weikang Wan, Jianyu Chen, Yuxiang Gao, and Mengdi Xu.
\newblock Hilo-wam: Decoupling long-horizon task evolution and local dynamics in world action models.
\newblock \emph{arXiv preprint arXiv:2611.05023}, 2026{\natexlab{c}}.

\bibitem[Wang et~al.(2024)Wang, Ma, and Cai]{templora}
Y. Wang, D. Ma, and D. Cai.
\newblock With greater text comes greater necessity: Inference-time training helps long text generation.
\newblock \emph{arXiv preprint arXiv:2401.11504}, 2024.

\bibitem[Wang et~al.(2026{\natexlab{d}})Wang, Huang, Li, Zhang, Liang, Jin, Chen, Chi, Zhou, Yu, et~al.]{openwam}
Yuran Wang, Siqiao Huang, Mingleyang Li, Chenhao Zhang, Jiaqi Liang, Weiyang Jin, Yue Chen, Xuemin Chi, Donghao Zhou, Qize Yu, et~al.
\newblock Openwam: An open, modular exploration towards systematic world-action model pretraining.
\newblock \emph{arXiv preprint arXiv:2609.07398}, 2026{\natexlab{d}}.

\bibitem[Wu et~al.(2023)Wu, Jing, Cheang, Chen, Xu, Li, Liu, Li, and Kong]{gr1bytedance}
Hongtao Wu, Ya Jing, Chilam Cheang, Guangzeng Chen, Jiafeng Xu, Xinghang Li, Minghuan Liu, Hang Li, and Tao Kong.
\newblock Unleashing large-scale video generative pre-training for visual robot manipulation.
\newblock \emph{arXiv preprint arXiv:2312.13139}, 2023.

\bibitem[Wu et~al.(2026)Wu, Shao, Qian, Zhao, Zhang, Zhao, Wang, Huang, Xu, Wang, et~al.]{recap}
Hao Wu, Tianyu Shao, Shengju Qian, Yizhou Zhao, Fan Zhang, Penghao Zhao, Weiyan Wang, Yufei Huang, Fan Xu, Kun Wang, et~al.
\newblock Remembering what matters for long-horizon robotic world models.
\newblock \emph{arXiv preprint arXiv:2610.14583}, 2026.

\bibitem[Yang et~al.(2026)Yang, Mu, Wei, Lu, Li, Xu, Xue, Yuan, Lin, Pang, and Xu]{memorywam}
Sizhe Yang, Juncheng Mu, Tianming Wei, Chenhao Lu, Xiaofan Li, Linning Xu, Zhengrong Xue, Zhecheng Yuan, Dahua Lin, Jiangmiao Pang, and Huazhe Xu.
\newblock Memorywam: Efficient world action modeling with persistent memory.
\newblock \emph{arXiv preprint arXiv:2606.20562}, 2026.

\bibitem[Yuan et~al.(2026{\natexlab{a}})Yuan, Qiao, Zhang, and Xu]{adaworldpolicy}
Ge Yuan, Qiyuan Qiao, Jing Zhang, and Dong Xu.
\newblock Adaworldpolicy: World-model-driven diffusion policy with online adaptive learning for robotic manipulation.
\newblock \emph{arXiv preprint arXiv:2602.20057}, 2026{\natexlab{a}}.

\bibitem[Yuan et~al.(2026{\natexlab{b}})Yuan, Dong, Liu, and Zhao]{fastwam}
Tianyuan Yuan, Zibin Dong, Yicheng Liu, and Hang Zhao.
\newblock Fast-wam: Do world action models need test-time future imagination?
\newblock \emph{arXiv preprint arXiv:2603.16666}, 2026{\natexlab{b}}.

\bibitem[Yuan et~al.(2026{\natexlab{c}})Yuan, Ji, Guan, Ye, Tian, Gu, Zhang, Yang, Huang, Li, et~al.]{cfwam}
Tingyu Yuan, Ziming Ji, Biaoliang Guan, Wen Ye, Wenrui Tian, Zhaopeng Gu, Feihong Zhang, Xu Yang, Yan Huang, Zhaowen Li, et~al.
\newblock From world models to world action models: Rethinking next-state prediction.
\newblock \emph{arXiv preprint arXiv:2609.34414}, 2026{\natexlab{c}}.

\bibitem[Zhang et~al.(2026)Zhang, Luo, Shi, Liu, Dong, Shan, Dai, Ren, and Ju]{geomemvla}
Qinfan Zhang, Yifan Luo, Haoyuan Shi, Guanlin Liu, Han Dong, Haozhe Shan, Yong Dai, Xiangcong Ren, and Xiaozhu Ju.
\newblock Geomem-vla: Internalizing 3d scene understanding into temporal memory for vision-language-action models.
\newblock \emph{arXiv preprint arXiv:2610.29866}, 2026.

\bibitem[Zhang et~al.(2025)Zhang, Bi, Hong, Zhang, Luan, Yang, Sunkavalli, Freeman, and Tan]{lact}
Tianyuan Zhang, Sai Bi, Yicong Hong, Kai Zhang, Fujun Luan, Songlin Yang, Kalyan Sunkavalli, William~T. Freeman, and Hao Tan.
\newblock Test-time training done right.
\newblock \emph{arXiv preprint arXiv:2505.23884}, 2025.

\bibitem[Zheng et~al.(2025)Zheng, Li, Wang, Liu, Kang, Feng, Zheng, Zou, Chen, Zeng, et~al.]{xvla}
Jinliang Zheng, Jianxiong Li, Zhihao Wang, Dongxiu Liu, Xirui Kang, Yuchun Feng, Yinan Zheng, Jiayin Zou, Yilun Chen, Jia Zeng, et~al.
\newblock X-vla: Soft-prompted transformer as scalable cross-embodiment vision-language-action model.
\newblock \emph{arXiv preprint arXiv:2510.10274}, 2025.

\bibitem[Zhou et~al.(2026)Zhou, Ye, Zhao, Dong, Wang, Ge, Yang, Hoorick, Sukhatme, Guizilini, et~al.]{rollingwam}
Yinghua Zhou, Junjie Ye, Yiqi Zhao, Hao Dong, Celina~Shiyu Wang, Ruohai Ge, Tingyi Yang, Basile~Van Hoorick, Gaurav Sukhatme, Vitor Guizilini, et~al.
\newblock Rolling-wam: World action models with rolling imagination.
\newblock \emph{arXiv preprint arXiv:2609.30247}, 2026.

\bibitem[Zhu et~al.(2025)Zhu, Yu, Feng, Burchfiel, Shah, and Gupta]{uwm}
Chuning Zhu, Raymond Yu, Siyuan Feng, Benjamin Burchfiel, Paarth Shah, and Abhishek Gupta.
\newblock Unified world models: Coupling video and action diffusion for pretraining on large robotic datasets.
\newblock \emph{arXiv preprint arXiv:2504.02792}, 2025.

\end{thebibliography}
